\pdfoutput=1
\documentclass[11pt]{article}
\usepackage[preprint]{acl}
\usepackage{times}
\usepackage{latexsym}
\usepackage[T1]{fontenc}
\usepackage[utf8]{inputenc}
\usepackage{microtype}
\usepackage{inconsolata}
\usepackage{graphicx}

\usepackage{hyperref}       
\usepackage{url}            
\usepackage{booktabs}       
\usepackage{amsfonts}       
\usepackage{nicefrac}       
\usepackage{xcolor}         
\usepackage{float}
\usepackage{booktabs}
\usepackage{array}
\usepackage{balance} 
\usepackage{multirow}
\usepackage[normalem]{ulem}
\usepackage{color}
\definecolor{lightgray}{RGB}{215,215,215}
\usepackage{colortbl}  
\usepackage{xcolor}
\useunder{\uline}{\ul}{}
\usepackage{subfigure}
\usepackage[utf8]{inputenc}
\usepackage{algorithm}
\usepackage{algpseudocode}
\usepackage{amsmath}  
\usepackage{enumitem}
\usepackage{tabularx}
\usepackage[utf8]{inputenc}
\usepackage[english]{babel}
\usepackage{amsthm}
\usepackage{bm}
\usepackage{graphicx}
\usepackage{caption}
\usepackage{wrapfig}
\usepackage{verbatim}

\usepackage{microtype}
\usepackage{graphicx}
\usepackage{booktabs} 
\usepackage{tcolorbox}
\usepackage{multicol}
\usepackage{cuted}
\usepackage{natbib}
\usepackage{graphicx}
\usepackage{booktabs}
\usepackage{amsmath}

\title{Safeguarding Vision-Language Models Against Patched Visual Prompt Injectors}

\author{
  \textbf{Jiachen Sun \textsuperscript{1}
  },
  \textbf{Changsheng Wang \textsuperscript{3}},
  \textbf{Jiongxiao Wang \textsuperscript{2}},
  \textbf{Yiwei Zhang \textsuperscript{2}},
  \textbf{Chaowei Xiao \textsuperscript{2}}
\\
\\
  \textsuperscript{1}University of Michigan Ann arbor,
  \textsuperscript{2}University of Wisconsin Madison,
\\
  \textsuperscript{3}Michigan State University,
\\ 
  \small{  
 jiachens@umich.edu, wangc168@msu.edu,
 $\{$jwang2929,zhang2486,cxiao34$\}$@wisc.edu
  }
}

\newtheorem{definition}{Definition}
\newtheorem{proposition}{Proposition}

\begin{document}
\maketitle

\begin{abstract}

Large language models have become increasingly prominent, also signaling a shift towards multimodality as the next frontier in artificial intelligence, where their embeddings are harnessed as prompts to generate textual content. Vision-language models (VLMs) stand at the forefront of this advancement, offering innovative ways to combine visual and textual data for enhanced understanding and interaction. However, this integration also enlarges the attack surface. Patch-based adversarial attack is considered the most realistic threat model in physical vision applications, as demonstrated in many existing literature. In this paper, we propose to address patched visual prompt injection, where adversaries exploit adversarial patches to generate target content in VLMs. Our investigation reveals that patched adversarial prompts exhibit sensitivity to pixel-wise randomization, a trait that remains robust even against adaptive attacks designed to counteract such defenses. Leveraging this insight, we introduce SmoothVLM, a defense mechanism rooted in smoothing techniques, specifically tailored to protect VLMs from the threat of patched visual prompt injectors. Our framework significantly lowers the attack success rate to a range between 0\% and 5.0\% on two leading VLMs, while achieving around 67.3\% to 95.0\% context recovery of the benign images, demonstrating a balance between security and usability.

\end{abstract}

\section{Introduction}

With the advent of large language models (LLMs) such as GPT and Claude~\cite{achiam2023gpt}, we have witnessed a transformative wave across numerous domains, guiding in an era where artificial intelligence (AI) closely mirrors human-like understanding and generation of language. This progress has further paved the way for the integration of multi-modality. Among them, vision-language models (VLMs)~\cite{zhang2024vision,chen2023pali} are emerging, which blend visual understanding with textual interpretation, offering richer interactions. However, as these VLMs grow more sophisticated, they also become targets for a wider range of adversarial threats. Attacks that involve altered visual prompts pose significant concerns, as they manipulate the models' responses in realistic ways that are hard to mitigate.

Many alignment studies focusing on LLMs appear to mitigate the spread of harmful content significantly ~\citep{ouyang2022training,bai2022constitutional,go2023aligning, korbak2023pretraining}. However, recent studies have exposed several vulnerabilities, known as \textit{jailbreaks}~\citep{chao2023jailbreaking}, which circumvent the safety measures in place for contemporary LLMs. Identifying and addressing these weaknesses presents significant challenges. They stand as major obstacles to the wider adoption and safe deployment of LLMs, impacting their utility across various applications. The integration of visual prompts arguably further enlarges the attack surface, introducing an additional layer of complexity for securing these systems. As models increasingly interpret and generate content based on both texts and images, the potential for exploitation through visually manipulated inputs escalates. This expansion not only necessitates advanced defensive strategies to safeguard against such innovative attacks but also underscores the urgent need for ongoing research and development in AI safety measures.

Although a variety of research has been conducted to study the robustness of jailbreak robustness of LLMs, there is a lack of practical formulation of ``visual jailbreaks'' as the emergence of VLMs. We thus first rigorously transform the existing adversarial attacks in VLMs~\citep{zhu2023minigpt,liu2023improved} as patched visual prompt injectors since patch-based attacks have been demonstrated as the most realistic attacks in the physical world.  As the ultimate goal of VLMs is text generation, the attack formulation is different from classic vision tasks such as 
classification~\citep{kri2017img} and object detection~\citep{zhao2019object} that target one-time logit outputs. There are two types of adversarial attacks for VLM that are prominent. 
\citep{shayegani2023jailbreak} propose to optimize the input visual prompt to mimic the harmful image in the embedding space, while \citep{qi2023visual} directly optimize the visual prompt to generate a given harmful content, as detailed in \S~\ref{sec:patched_visual_prompt_injection}. We adopt both optimization methods but update the attack interface from $\ell_\infty$-bounded manipulations to adversarial patches.   
This vulnerability not only undermines the reliability of these systems but also poses significant security risks, especially in critical applications. The need to safeguard against such vulnerabilities is not just imperative for the integrity of VLMs but is also of paramount importance for the trust and widespread adoption of LLMs and VLMs.

In this paper, we further introduce SmoothVLM, a novel framework designed to fortify VLMs against the adversarial threat of patched visual prompt injectors. SmoothVLM is designed to naturally enhance the robustness against visual jailbreaks while preserving the interpretative and interactive performance of VLM agents. We first identify an intriguing property of the patched visual prompt injectors, that is, the success of the injection is extremely sensitive to the random perturbation of the adversarial patch even under adaptive attacks. This could be attributed to the design of VLM. Therefore, by integrating majority voting with random perturbed visual prompts, our approach can defend the hidden visual prompt injectors with high probability, effectively rendering them impotent in manipulating model behavior. SmoothVLM has significantly reduced the attack success rates of patched visual prompt injectors on popular VLMs. Specifically, for both \texttt{llava-1.5} and \texttt{miniGPT4}, SmoothVLM can reduce the attack success rate (ASR) to below 5\%, and with a sufficiently large perturbation, it can further decrease the ASR to approximately 0\%.

Our contributions are manifold and significant:

$\bullet$ We present a comprehensive analysis of the vulnerabilities of current VLMs to patched visual prompt attacks and propose SmoothVLM, a novel defense mechanism that leverages randomized smoothing to mitigate the effects of adversarial patches in VLM.

$\bullet$ We demonstrate through extensive experiments that SmoothVLM significantly outperforms existing defense strategies, achieving state-of-the-art results in both detection accuracy and model performance retention.

$\bullet$ By addressing the susceptibility of VLMs to adversarial patch-based manipulations even under adaptive attacks, SmoothVLM represents a significant step forward in the development of secure multimodal LLMs.
\section{Related Work}
\label{sec:relatedwork}

In this section, we review a few related topics to our study, including attacks and defenses for prompt injection and adversarial patches.

\subsection{Prompt Injection}

Prompt engineering is emerging in the era of LLM. At the core of prompt injection attacks lies the adversarial ability to manipulate the output of LLMs by ingeniously crafting input prompts. \citep{zhang2020} provided an early exploration of these vulnerabilities in LLMs, demonstrating how attackers could insert malicious prompts to alter the behavior of AI systems in text generation tasks. Their work highlighted the need for robust input validation and sanitization mechanisms to mitigate such threats.  \cite{zou2023universal} conducted an empirical study on the impact of prompt injection attacks on various commercial AI systems, uncovering a wide range of potential exploits, from privacy breaches to the spread of misinformation. Recently, prompt injection attacks extended to VLMs~\cite{bailey2023image}. In particular, \cite{shayegani2023jailbreak} and~\cite{qi2023visual} propose to modify the pixels of the visual prompts to fool VLMs that generate target contents, as detailed in \S~\ref{sec:patched_visual_prompt_injection}.

\begin{figure*}
    \centering
    \vspace{-0.5cm}
    \includegraphics[width=0.9\linewidth]{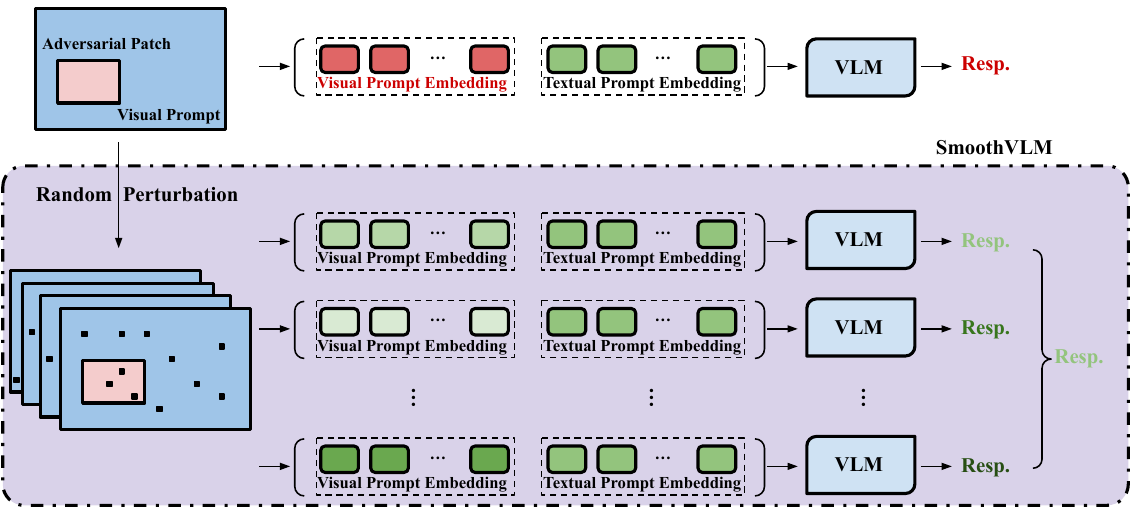}
    \vspace{-0.2cm}
    \caption{\footnotesize \textbf{Our SmoothVLM Certified Defense Pipeline.}}
    \label{fig:pipeline}
    \vspace{-0.5cm}
\end{figure*}

\subsection{Adversarial Patches}

The advent of adversarial patch attacks has prompted significant research interest due to their practical implications for the security of machine learning systems, especially those relying on computer vision. This section reviews key contributions to the field, spanning the initial discovery of such vulnerabilities to the latest mitigation strategies. ~\citep{46561} pioneered the exploration of adversarial patches by demonstrating that strategically designed and placed stickers could deceive an image classifier into misidentifying objects. Following the initial discovery, researchers sought to refine the techniques for generating and deploying adversarial patches. ~\cite{AM_2015} introduced an optimization-based method to create more effective and efficient adversarial patches.  The practical implications of adversarial patch attacks have been a focus of recent studies. ~\cite{chahe2023dynamic} investigated the effects of adversarial patches on autonomous vehicle systems, revealing potential threats to pedestrian detection mechanisms. In response to these vulnerabilities, the community has developed various defensive strategies. ~\citep{strack2023defending} proposed a defense mechanism based on anomaly detection and segmentation techniques to identify and ignore adversarial patches in images. ~\cite{zhou2021towards} explored the integration of adversarial examples, including patches, into the training process. Certified defenses also extend to adversarial patches. ~\citet{xiang2022patchcleanser, xiang2024patchcure} developed certified methods to mitigate adversarial patches. However, the current certified methods are only limited to defending against small adversarial patches.

\section{SmoothVLM}

In this section, we present our SmoothVLM framework to defend against adversarial patches for visual prompt injection. We first introduce our threat model of patched visual prompt injection.

\subsection{Patched Visual Prompt Injection}
\label{sec:patched_visual_prompt_injection}
We have witnessed the emergence and potential of large (vision) language models (LLM and VLM) in the past year and they have also introduced new attack vectors such as prompt injection~\citep{liu2023prompt,greshake2023not, shi2024optimization}. Different from classic adversarial attacks targeting fundamental tasks such as classification and object detection that target logit space manipulation, prompt injection aims to induce language models to generate specific texts. A VLM incorporates multimodality by treating images as visual prompts to an appended LM, enhancing the model's comprehension of instructions. To inject a target concept into the VLM, there are currently two primary optimization methods. Firstly, \citep{shayegani2023jailbreak} optimized the distance between embeddings of the adversarial and target images (\textit{e.g.}, a bomb or a gun), \textit{i.e.}, $\arg \min_{x_\text{adv}} d(H_\text{adv}, H_\text{target})$, where $H$ denotes the visual embedding ingested by the LM, ensuring the LM cannot discern between adversarial and target image embeddings as long as the distance $d(H_\text{adv}, H_\text{target})$ is minimal.
Secondly, \cite{qi2023visual} proposed using a corpus of harmful text as the target to optimize the image input, \textit{i.e.}, $\arg \min_{x_\text{adv}} \sum_{i=1}^m - \log(p(y_i|[x_\text{adv},\emptyset]))$, where $Y_\text{adv}:=\{y_i\}_{i=1}^{m}$ represents the corpus of chosen content. Both studies leverage the $\ell_\infty$ norm across the image's pixel attack surface. However, DiffPure~\cite{nie2022diffusion} and its follow-ups have shown that such threat models can be mitigated through diffusion purification, with many subsequent studies~\cite{wang2023better,lee2023robust,zhang2023diffsmooth,xiao2022densepure} confirming its effectiveness against both $\ell_2$ and $\ell_\infty$-based attacks. Therefore, we propose adapting these two attack strategies to use adversarial patches with an $\ell_0$ constraint on size, which both maintains the stealthiness of the attack and the original image's semantics. Patch attacks are also demonstrated to be much more physically achievable in the real world. We denote the two attack methods by their titles Jailbreak In Pieces (JIP) and Visual Adversarial Examples (VAE), respectively.

Specifically, our threat model assumes that an adversarial patch $P_{\text{m}\times\text{n}}^{[i,j]}$, of size $\text{m}\times\text{n}$, is placed such that its bottom-left corner aligns with the pixel at coordinates $[i,j]$ in the original image $I_{\text{h}\times\text{w}}$. Here, $I_{\text{h}\times\text{w}}$ denotes the original image of size $\text{h}\times\text{w}$. The resultant adversarial example is denoted as $I_\text{adv} = I \oplus P$, where $\oplus$ signifies the operation used to overlay the patch onto the image. We still leverage the two white-box optimization methods mentioned above in our evaluation.

\subsection{Randomized Defense Against Patched Visual Prompt Injection}
\label{sec:q_unstable}
As introduced in \S~\ref{sec:relatedwork}, randomized defenses are significant within the adversarial robustness community. Drawing inspiration from SmoothLLM~\citep{robey2023smoothllm}, our investigations reveal vulnerabilities to randomized perturbations in the pixel space of the patched visual prompt injectors. In preliminary experiments on the latest \texttt{LLaVA-v1.5-13b} model~\citep{liu2024visual,liu2023improved}, which accepts $224 \times 224$ images. Since adversarial optimization is computationally expensive, we leverage 300 adversarial examples and ensure that the attacks successfully launch on the images. We applied three randomized perturbation methods to the adversarial patch area in the images: \textit{mask}, \textit{swap}, and \textit{replace}. The \textit{mask} operation randomly sets $q$\% of the pixels in the adversarial patch to zero across all channels. For \textit{swap}, $q$\% of the pixels' RGB channels are randomly interchanged. The \textit{replace} operation substitutes $q$\% pixels with random RGB values uniformly sampled.

As mentioned earlier, JIP and VAE are both optimized to generate $Y_\text{adv}$ or its equivalents.
Similar to~\citet{robey2023smoothllm}, we leverage an oracle language model (e.g., GPT4~\citep{achiam2023gpt}) to deterministically predict whether the attack goal is achieved. Therefore, a successful attack (SA) is defined as:  
\begin{align*}
    \text{VPI}(Y_\text{pred})&\doteq \text{OracleLLM}(Y_\text{pred}, Y_\text{adv})\\
    & = \begin{cases}
      1, & \text{if}~Y_\text{pred},~Y_\text{adv}~\text{are synonymous}  \\
      0, & \text{otherwise.}
    \end{cases}
\end{align*}
Figure~\ref{fig:q_instability} shows the attack success rates (ASR) of our preliminary measurement study, which illustrate the 
instability of the patched visual prompt injection attacks. We found that among the three types of perturbation, random masking can consistently and effectively mitigate adversarial patch attacks with a sufficient amount of perturbation. 
Particularly, random masking reduces the ASR to around 5\%. 
We denote our finding as \textbf{visual q-instability with probability error $\epsilon$}.

\subsection{Expectation over Transformation (EOT) Adversary} 
\label{sec:eot}
\begin{figure*}[t]
 \vspace{-0.5cm}
  \includegraphics[width=0.48\linewidth]{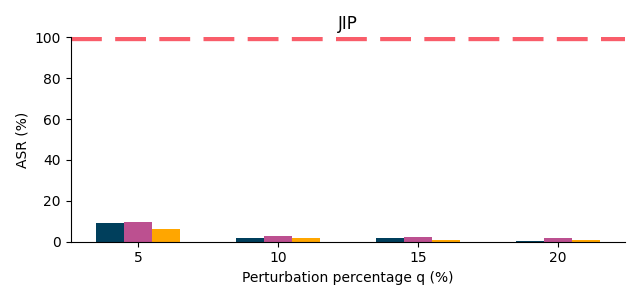} \hfill
  \includegraphics[width=0.48\linewidth]{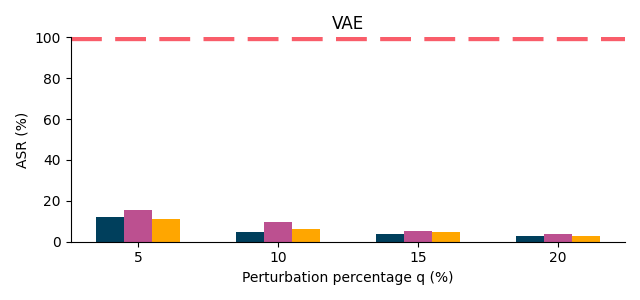}
       \vspace{-0.2cm}
  \caption{\footnotesize \textbf{Validation of \textit{q}-instability on Patched Visual Prompt Injection.} We random perturb $q$\% pixels in the adversarial patch with three methods: \textit{mask}, \textit{swap}, and \textit{replace}. The red dashed line shows the ASR of the attack method JIP and VAE.}
    \label{fig:q_instability}
     \vspace{-0.5cm}
\end{figure*}
\begin{figure*}[t]
\vspace{-0.5cm}
  \includegraphics[width=0.45\linewidth,height=4cm]{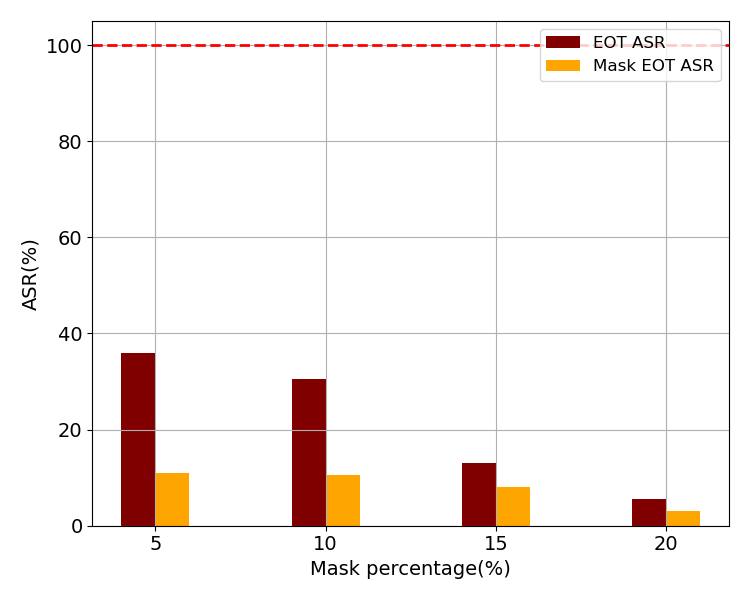} \hfill
  \includegraphics[width=0.45\linewidth,height=4cm]{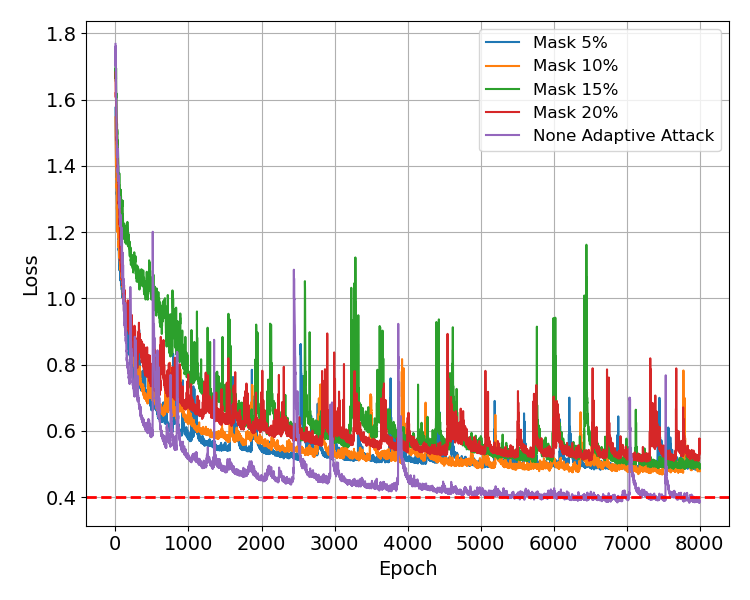}
  \vspace{-0.2cm}
  \caption{\footnotesize \textbf{Validation of q-instability on EOT Attack.} The left figure plots the ASR of EOT adversarial examples w/wo \textit{q}\% pixels masked. The red dashed line at the ASR of 100\% denotes that all the original samples are successfully attacked. ``Mask EOT ASR'' means that after we get adversarial examples with EOT, we further mask \textit{q}\% pixels as our defense. 
  For the right subplot, we plot the training loss with 8000 epochs (requiring $\sim$50 mins on one A100), ``mask \textit{q}\%'' means we mask \textit{q}\% in EOT attack process, ``none adaptive attack'' means normal patch attack. The dotted red line in the right figure indicates the required loss for a successful adversarial optimization, \textit{i.e.}, loss=0.4. \textbf{The two figures demonstrate that EOT is extremely hard to optimize and subject to our identified \textit{q}-instability as well.}}
  \vspace{-0.3cm}
    \label{fig:eot}
\end{figure*}

Randomization-based defense solutions can be oftentimes broken by the expectation of transformations (EOT) attack~\citep{athalye2018synthesizing} in the existing literature. However, we argue that adaptive attacks are more challenging to launch in the era of multimodal language models, especially under realistic threat models. We empirically demonstrate that the EOT attacks are ineffective in breaking the q-instability of VLM under our patched visual prompt injection setup. Specifically, we assume the attacker is aware of the exact random perturbation (masking here since it is demonstrated to be the most effective one in Figure~\ref{fig:q_instability}) added for defense; thus, the optimization becomes
\begin{equation}
    {\arg \min}_{P} \mathbb{E}_{t \sim {Mask()}} \ell(\text{VLM}([I \oplus t(P);\emptyset]), Y_{adv})
\end{equation}
where $t$ follows the distribution of our random masking. 
As shown in Figure~\ref{fig:eot}, the ASRs of the adaptive attacks are extremely low. 
\begin{tcolorbox}[left=1mm, right=1mm, top=0mm, bottom=0mm]
\footnotesize
\begin{definition}
\label{def:kunstable_1}
\textbf{Definition 1 (Visual \textit{q}-instability with probability error $\epsilon$).} Given a VLM and the adversarial example $I^{\text{adv}}=I \oplus P$, we apply the \textit{mask} operation Mask() to randomly zero out $q$\% pixels in the adversarial patch $P$, obtaining $P' = \text{Mask}(P)$. Here we call the $P$ is \textbf{visual \textit{q}-unstable with probability error $\epsilon$} if for any
\begin{equation}
    \ell_0(P',P) \geq \ulcorner qmn \urcorner
\end{equation}
there exists a small constant probability error $\epsilon$ such that 
\begin{equation}
    Pr[(\text{VPI} \circ \text{VLM}) ([I \oplus P';\emptyset]) = 0] \geq 1-\epsilon
\end{equation}
where $\textbf{\textit{q}}$ is the instability parameter.
\end{definition}
\end{tcolorbox}

The reason could be attributed to the characteristics of the VLM task, which is essentially the \textbf{\textit{next-token prediction}}. 
As introduced earlier, the attack goal for classic vision tasks is to manipulate a single/few output(s) from the one-time model inference, so the room for adversarial optimization is arguably large. However, the optimization goal is either too harsh or implicit for next-token prediction as it usually involves a sequence of outputs with many iterations.

Figure~\ref{fig:eot} that is hard to optimize. For example, the loss function for JIP is the mean square error, the $\ell_2$ distance in the embedding space. In LLaVA, there are 576 token embeddings, and a successful attack needs to make the $\ell_2$ between less than 0.4, which is far more difficult than the classic logit space optimization when combined with randomized defense. In the VAE, the optimization directly targets generating harmful content. As there are usually 8000 iterations, which already take 0.5 hours to optimize, EOT will make the complexity at least an order higher, rendering the optimization intractable. 

\subsection{SmoothVLM Design}

We formally introduce our design of SmoothVLM. Similar to SmoothLLM and other randomized smoothing-based methods, there are two key components: (1) distribution procedure, in which $N$ copies of the input image with random masking are distributed to VLM agents for parallel computing, and (2) aggregation procedure, in which the responses corresponding to each of the perturbed copies are collected, as depicted in Figure~\ref{fig:pipeline}.

\subsubsection{Distribution Procedure}
The first step in our SmoothVLM is to distribute $N$ visual prompts to the VLM agent and this step can be computed in parallel if resources are allowable.
As illustrated in \S~\ref{sec:q_unstable}, when the patched visual prompt injectors $P$ are \textbf{\textit{q}-unstable with probability error $\epsilon$}, the probability of a successful defense is no lower than 1-$\epsilon$. However, in most situations, we do not know where the adversarial patch is attached to the prompt ( $i,j$), so we can only apply random perturbation to the whole visual prompt. Shown as following Assumption 2, here we assume that the masking pixels out of adversarial patches would at most lead to a decreasing of $\mu$ on the probability of a successful defense. 
Since masking is the most stable perturbation shown in Figure~\ref{fig:q_instability}, we will use masking in the rest of this paper. Specifically, we randomly mask $q\%$ of the pixels for each distributed visual prompt.

\begin{tcolorbox}[left=1mm, right=1mm, top=0mm, bottom=0mm]
\footnotesize
\begin{definition}
\label{def:kunstable_2}
\textbf{Assumption 2 (Visual \textit{q}-instability for Visual Prompt $I^{adv}$).} Given a VLM and the adversarial example $I^{\text{adv}} = I \oplus P$, we apply the \textit{mask} operation Mask() to randomly zero out $q$\% pixels in the visual prompt $I^{\text{adv}}$. Here we denote $I'=\text{Mask}(I)$, $P'=\text{Mask}(I)_{|P}$,  which means then projection of the $\text{Mask}(I)$ on the position of the adversarial patch P. If P is visual $q$-unstable with probability error $\epsilon$, we assume that 
\begin{equation}
Pr[(\text{VPI} \circ \text{VLM}) ([I' \oplus P';\emptyset]) = 0] \geq 1-\epsilon-\mu
\end{equation}
The consideration of $I'$ instead of $I$ would at most lead to a decreasing of $\mu$ in the probability of a successful defense.
\end{definition}
\end{tcolorbox}

\subsubsection{Aggregation Procedure}

The second step in our SmoothVLM is to collect and aggregate the responses from the first step. 
As mentioned earlier, as we do not know the location of the adversarial patch, it is impossible to guarantee high defense probability with one masked input. Therefore, rather than passing a single perturbed prompt
through the LLM, we obtain a collection of perturbed prompts with the same perturbation rate p, and then aggregate the predictions
of this collection. The motivation for this step is that while one perturbed prompt may not
mitigate an attack, as we observed in Figure 4, on average, perturbed prompts tend to nullify jailbreaks. That
is, by perturbing multiple copies of each prompt, we rely on the fact that on average, we are likely to flip
characters in the adversarially-generated portion of the prompt. 

Based on the above two insights, here we give the formal definition of SmoothVLM in Definition 3 and include the details about the algorithm in Algorithm 1.

\subsection{Probability Guarantee of SmoothVLM}
To understand the robustness of SmoothVLM against VPI, we also provided a thorough analysis of the defense success probability (DSP) $\text{DSP}([I;\emptyset])$. Here we give the result of DSP in Proposition 4. Detailed computation process is included in Appendix~\ref{appendix:proof}.

\begin{algorithm}
\footnotesize
\caption{SmoothVLM}
\begin{algorithmic}[1]
\Require Visual Prompt $I$
\State \textbf{Input:} Number of Samples $N$, Perturbation Rate $p$
\For{$j = 1 \dots N$}
    \State $I_j \gets \text{RandomPerturbation}(I, p)$
    \State $R_j \gets \text{VLM}([I_j;\emptyset])$
\EndFor
\State $A \gets \text{MajorityVote}(R_1, \dots, R_j)$
\State $j^* \sim \text{Unif}(\{j \in [N] \: | \: R_j = A\})$
\State \textbf{return} $R_{j^*}$
\Statex
\Function{MajorityVote}{$R_1, \dots, R_N$}:
    \State \textbf{return} $\mathbb{I} \biggl[ \frac{1}{N} \sum_{j=1}^N \text{VPI}(R_j) \geq \frac{1}{2} \biggr]$
\EndFunction
\end{algorithmic}
\end{algorithm}

\begin{tcolorbox}[left=1mm, right=1mm, top=0mm, bottom=0mm]
\footnotesize
\begin{definition}
\label{def:inj}
\textbf{Definition 3 (SmoothVLM).} Let a visual prompt (image) $I$ and a distribution $\mathbb{P}_p(I)$ over randomly masked copies of $I$ be given. Let $I_1, ..., I_{N}$ be drawn i.i.d. from $\mathbb{P}_p(I)$
\begin{equation}
    A \doteq \mathbb{I}[\frac{1}{N}\sum_{j=1}^{N}(\text{VPI} \circ \text{VLM})(I_j) > \frac{1}{2}]
\end{equation}
and our SmoothVLM is defined as
\begin{equation}
    \text{SmoothVLM}([I;\emptyset]) \doteq \text{VLM}([\mathbf{I};\emptyset])
\end{equation}
where $\mathbf{I}$ represents the image agrees with majority voting, $(\text{VPI} \circ \text{VLM})([\mathbf{I};\emptyset]) = A$.
\end{definition}
\end{tcolorbox}

\begin{figure*}[t]
\vspace{-0.5cm}
  \includegraphics[width=0.32\linewidth]{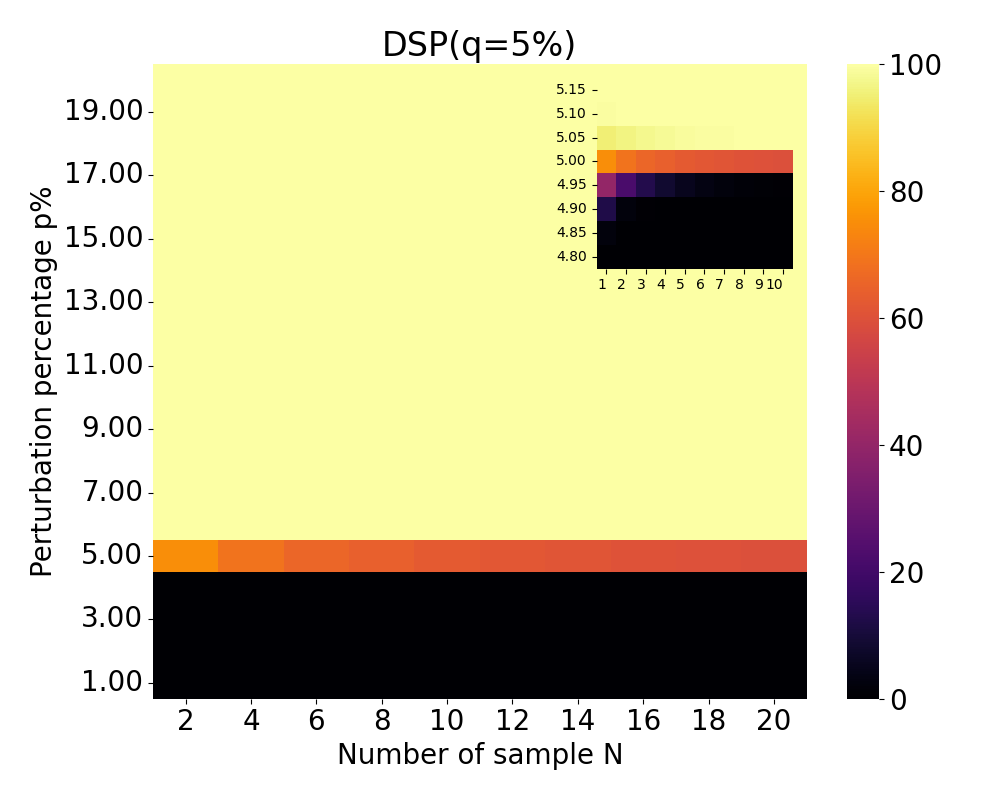} \hfill
  \includegraphics[width=0.32\linewidth]{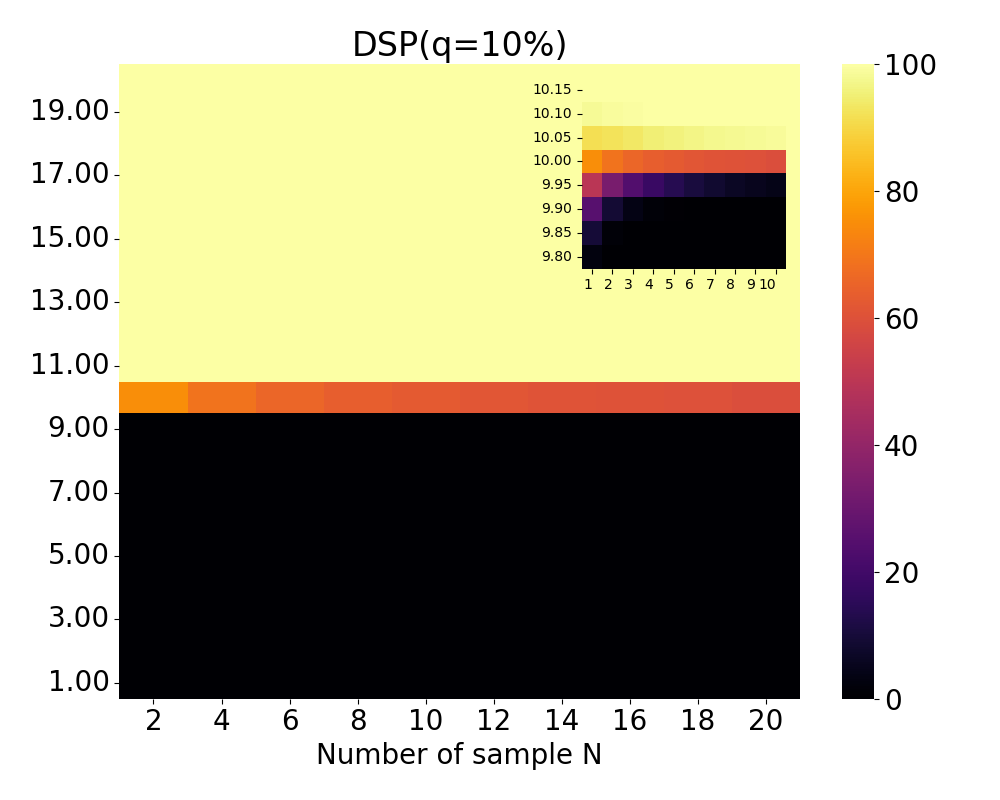} \hfill
  \includegraphics[width=0.32\linewidth]{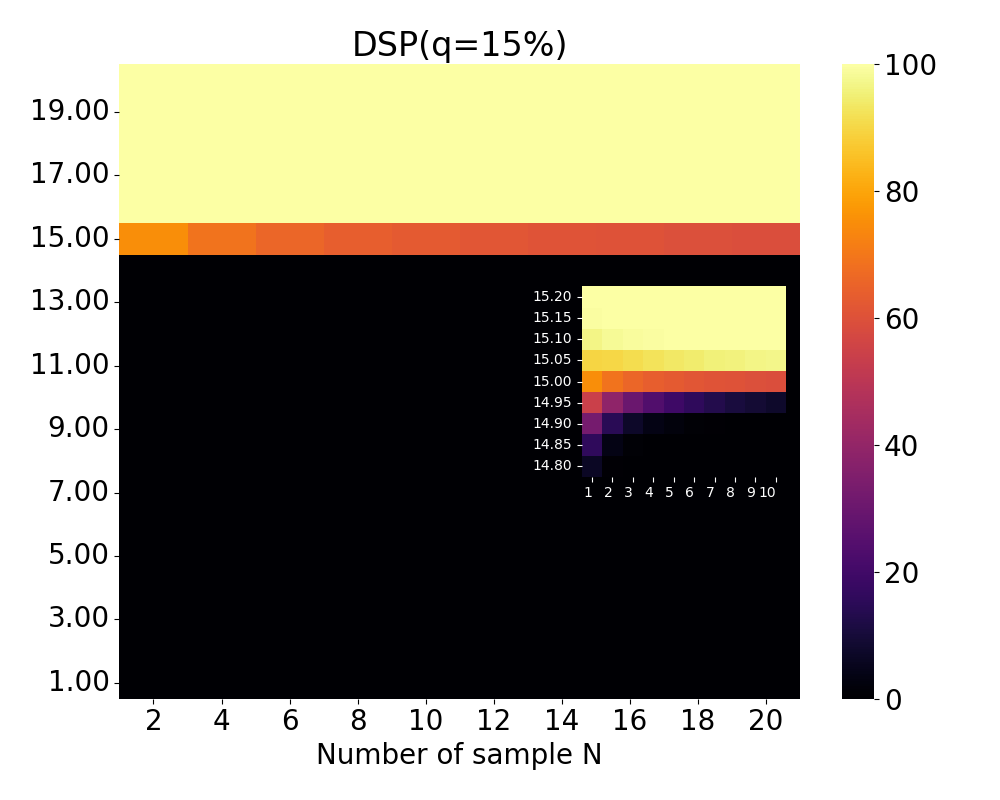} \\
    \vspace{-0.7cm}
  \caption{\footnotesize \textbf{Robustness Guarantee on Patched Visual Prompt Injection.} We plot the probability $\text{DSP}([I \oplus P;\emptyset])$ that SmoothVLM will consider attacks as a function of the number of samples $N$ and the perturbation percentage $q$; warmer colors denote larger probabilities. From left to right, probabilities are calculated for ten distinct values of the instability parameter $k$ from 2 to 20. Each subplot reveals the pattern: with the increase in both $N$ and $q$, there is an increasing DSP. }
  \label{fig:heatmap}
  \vspace{-0.1cm}
\end{figure*}


\begin{figure*}
\begin{tcolorbox}[left=1mm, right=1mm, top=1mm, bottom=1mm]
\footnotesize
\begin{proposition}
\textbf{Proposition 4 (Defense Success Probability of SmoothVLM).} Assume that an adversarial patch $P\in [0,1]^{\text{m}\times\text{n}\times3}$ for the visual prompt $I_{\text{h}\times\text{w}} \in [0,1]^{\text{h}\times\text{w}\times3}$
is \textbf{visual \textit{q}-unstable with probability error $\epsilon$}. Recall that $N$ is the number of randomly masked samples drawn i.i.d. and $p$ is the perturbation percentage on the whole visual prompt. The DSP is derived as follows:
\begin{align}
 \text{DSP}([I \oplus P;\emptyset]) & = Pr[(\text{VPI} \circ \text{SmoothVLM})([I \oplus P;\emptyset])=0]  \\
& = \sum_{t=\ulcorner N/2 \urcorner}^N \binom{N}{t}\alpha^{t}(1-\alpha)^{N-t} \\
\text{where} \quad \alpha & \geq (1-\epsilon-\mu)\sum_{k=\ulcorner qmn \urcorner}^{mn} \binom{mn}{k}p^{k}(1-p)^{mn - k}
\end{align}
\end{proposition}
\end{tcolorbox}
\end{figure*}


\section{Evaluations}

\begin{figure*}[!t]
\vspace{-0.4cm}
  \centering
  \includegraphics[width=0.3\linewidth, height=3cm]{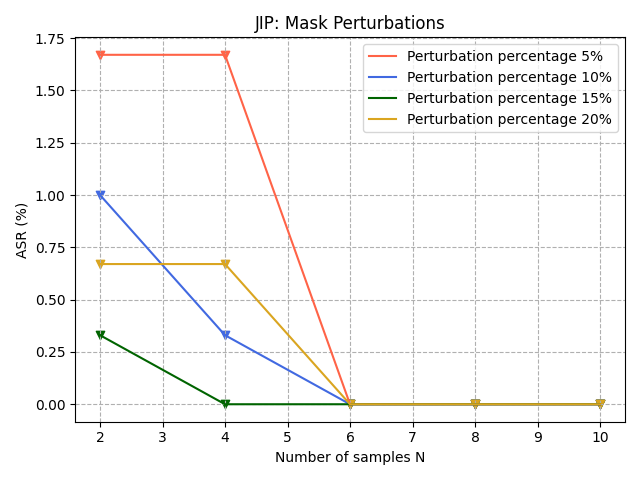} 
  \includegraphics[width=0.3\linewidth, height=3cm]{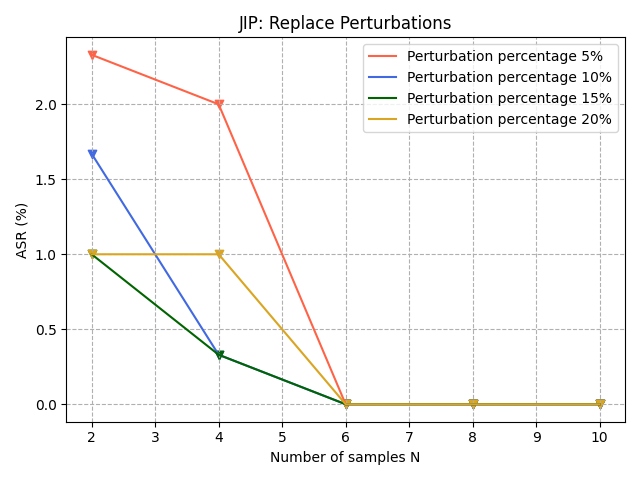} 
  \includegraphics[width=0.3\linewidth, height=3cm]{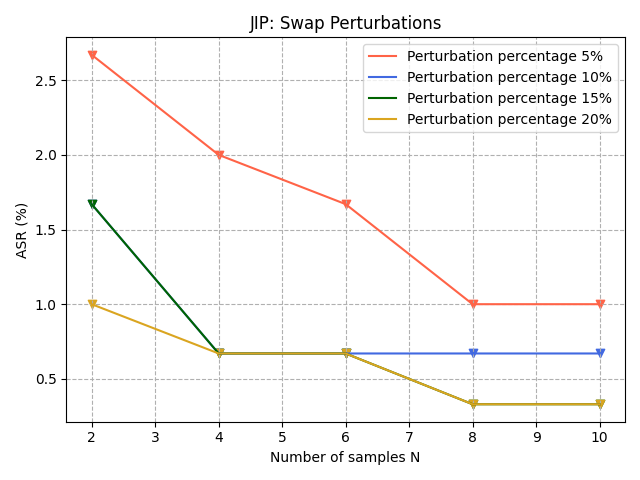} \\
  \includegraphics[width=0.3\linewidth, height=3cm]{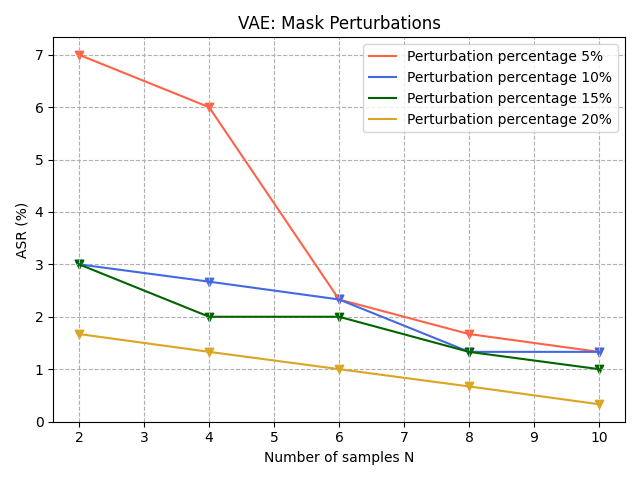} 
  \includegraphics[width=0.3\linewidth, height=3cm]{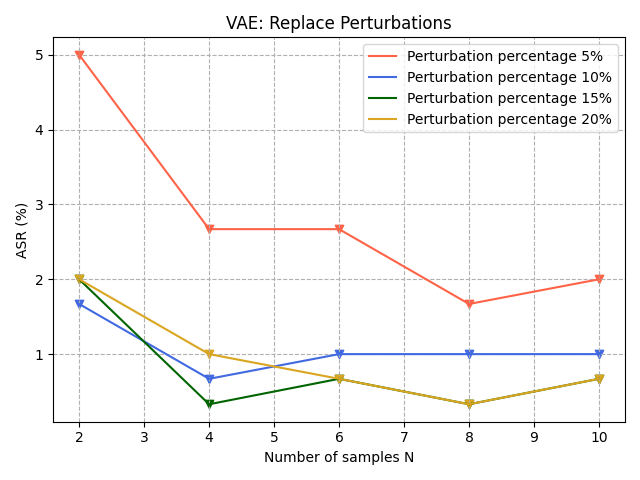} 
  \includegraphics[width=0.3\linewidth, height=3cm]{ 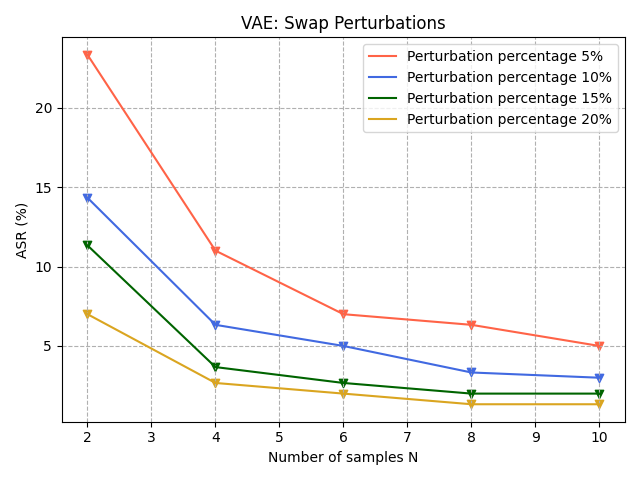}
  \vspace{-0.2cm}
   \caption{\footnotesize \textbf{Injection Mitigation Effectiveness of SmoothVLM.} We plot the ASR of VLM patch attack JIP (top row) and VAE (bottom row) for various values of the perturbation percentage $q \in \{5, 10, 15, 20 \}$ and the number of samples $N \in \{2, 4, 6, 8, 10 \}$.}
    \label{fig:main_tabel_attack}

      \centering
  \includegraphics[width=0.3\linewidth, height=3cm]{ 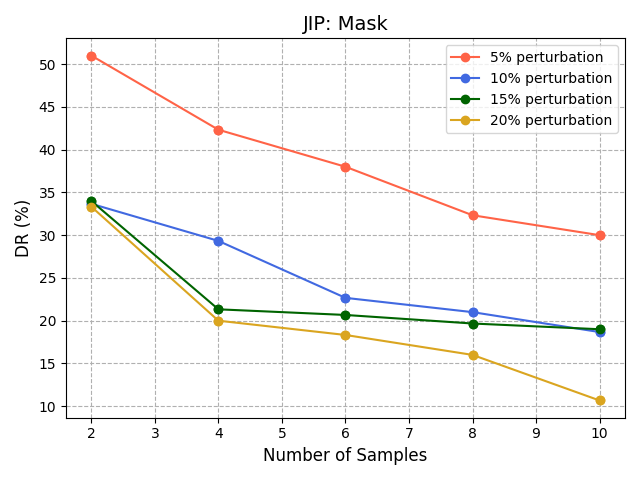} 
  \includegraphics[width=0.3\linewidth, height=3cm]{ 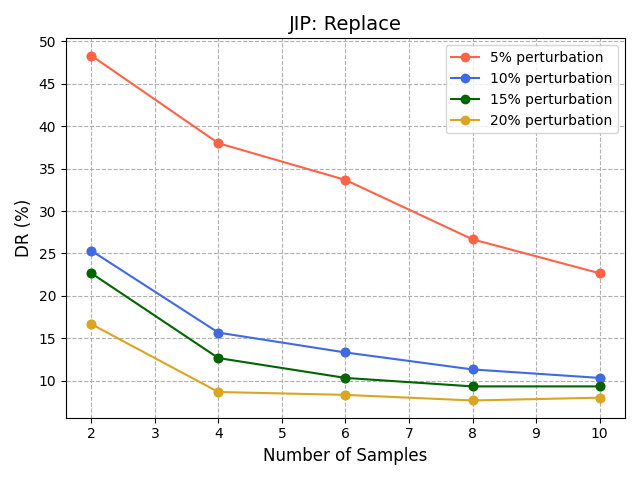} 
  \includegraphics[width=0.3\linewidth, height=3cm]{ 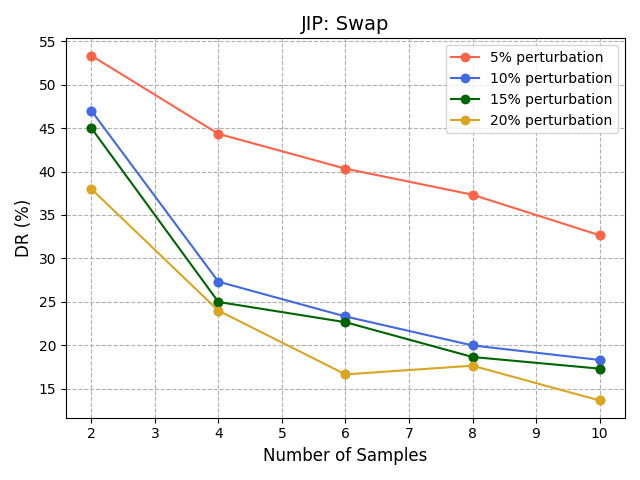} \\
  \includegraphics[width=0.3\linewidth, height=3cm]{ 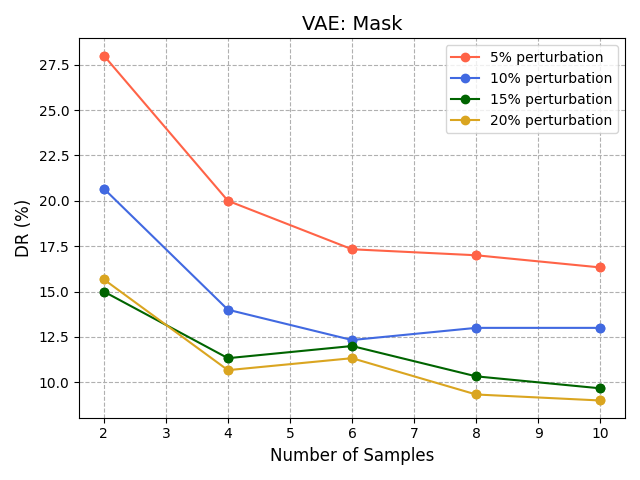} 
  \includegraphics[width=0.3\linewidth, height=3cm]{ 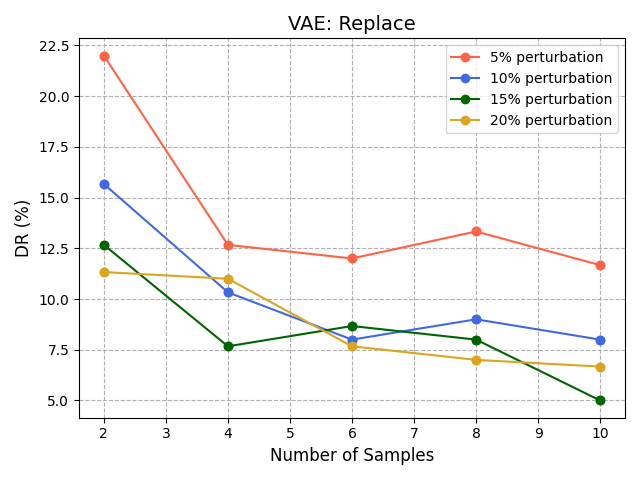} 
  \includegraphics[width=0.3\linewidth, height=3cm]{ 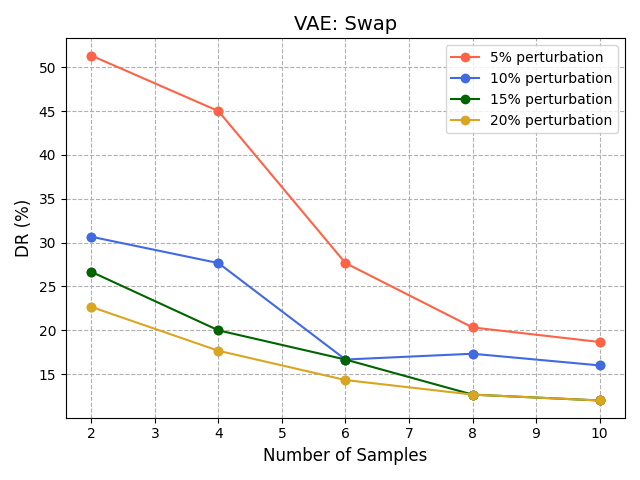}
    \vspace{-0.2cm}
   \caption{\footnotesize \textbf{Visual Prompt Recovery.} We plot the distortion rate of VLM patch attack JIP (top row) and VAE (bottom row) for various values of the perturbation percentage $q \in \{5, 10, 15, 20 \}$ and the number of samples $N \in \{2, 4, 6, 8, 10 \}$.}
    \label{fig:main_tabel_source}

      \centering
  \includegraphics[width=0.3\linewidth, height=3cm]{ 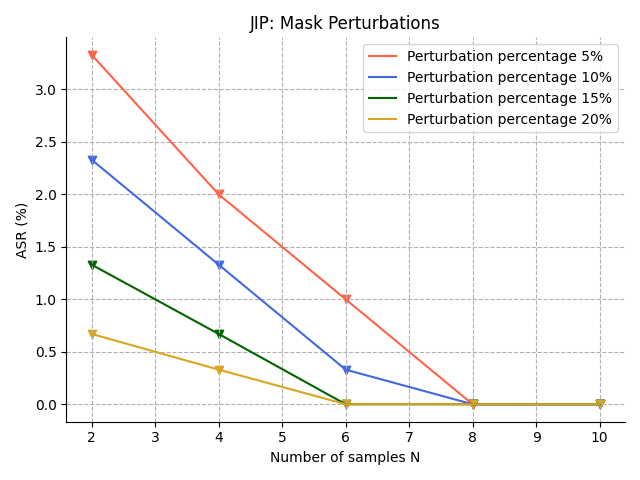} 
  \includegraphics[width=0.3\linewidth, height=3cm]{ 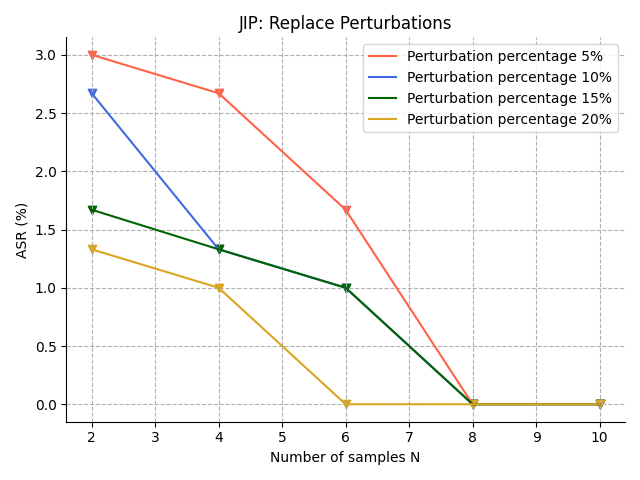} 
  \includegraphics[width=0.3\linewidth, height=3cm]{ 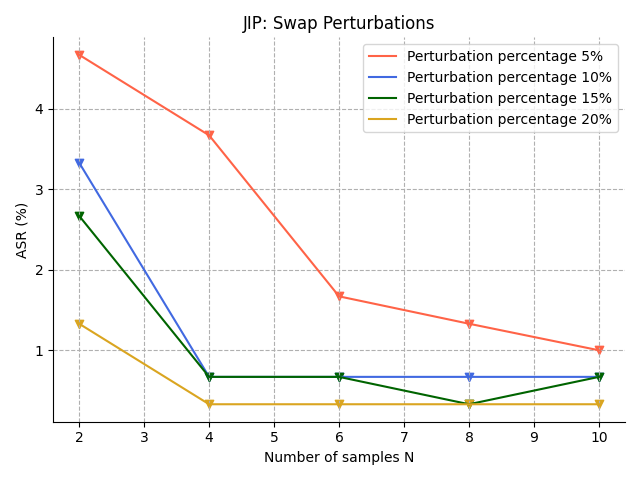} \\
  \includegraphics[width=0.3\linewidth, height=3cm]{ 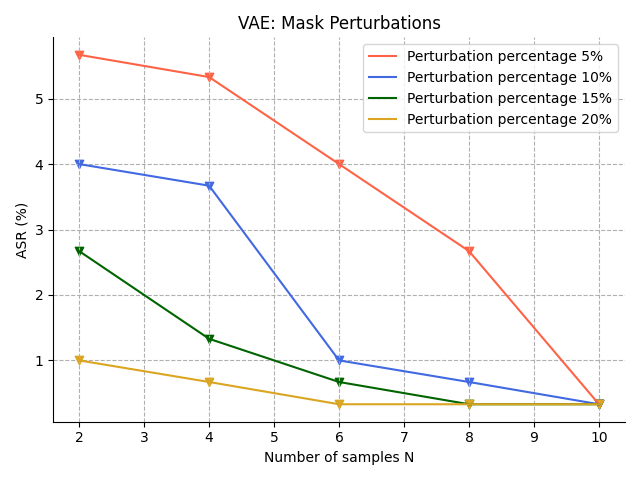} 
  \includegraphics[width=0.3\linewidth, height=3cm]{ 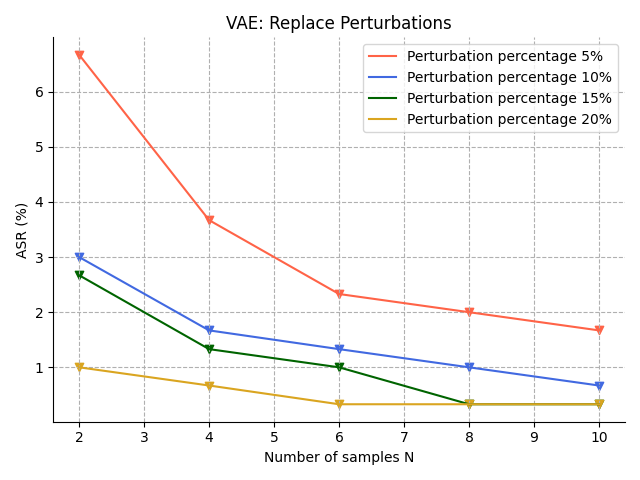} 
  \includegraphics[width=0.3\linewidth, height=3cm]{ 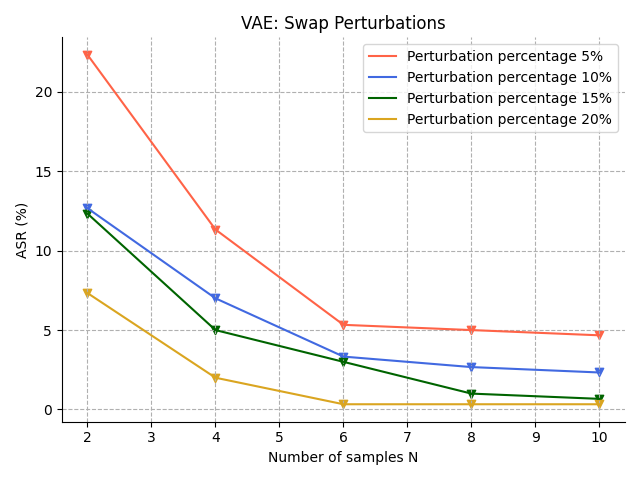}
    \vspace{-0.2cm}
   \caption{\footnotesize \textbf{Dual-Patched Injection Mitigation Effectiveness of SmoothVLM.} We plot the ASR of VLM patch attack JIP (top row) and VAE (bottom row) for various values of the perturbation percentage $q \in \{5, 10, 15, 20 \}$ and the number of samples $N \in \{2, 4, 6, 8, 10 \}$.}
    \label{fig:double_tabel_attack}
    \vspace{-0.4cm}
\end{figure*}

In this section, we conduct a comprehensive evaluation of our proposed SmoothVLM, which mainly show the results of two attack methods on \texttt{llava-1.5}. Specifically, we leverage \texttt{Vicuna-30B} as a proxy function for $VPI()$. 

\subsection{Injection Mitigation}

We conducted extended experiments on two attack methods on both \texttt{llava-1.5} and \texttt{miniGPT4}, mainly presents the results for \texttt{llava-1.5} in this section. Notably, we utilized 300 adversarial examples, all verified against the corresponding VLM model to ensure a successful attack. In Figure~\ref{fig:main_tabel_attack} displays various values of the number of samples $N$ and the perturbation percentage $q$. Generally, the attack success rate (ASR) significantly decreases as both q and N increase. Specifically, it's observed that even with a minimal perturbation $q$=5\%, increasing the sample number $N$ leads to a substantial drop in ASR. And when $q$=5\%, we can find that the ASR of various methods is significantly higher than at other percentage rates, especially when $N$ is also very small. For the three methods, we can clearly see that the ASR of the Swap method is significantly higher than that of mask and replace, which is consistent with the results at the $q$-instability point.




\subsection{Visual Prompt Recovery}

The goal is to recover the original semantics. We evaluate the similarity between the responses generated from the perturbed image and the original image to determine if the perturbation can recover the adversarial example to its original state. The distortion rate quantifies the discrepancy between the response to the original image and the response after perturbation.  Here we use \texttt{Vicuna-30B} as a metric function. In Figure, the small value of $q$ = 5\% results in a higher distortion rate, suggesting that lower perturbation levels are insufficient to eliminate the concealed harmful context within the visual prompt and fail to restore the original visual semantics.


\subsection{Efficiency}
Smoothing-based certified defenses unavoidably increase the inference runtime~\cite{xiang2022patchcleanser,cohen2019certified}.  In this section, we demonstrate that our SmoothVLM is efficient from two perspectives. We first compare the efficiency of the attack method (JIP) with our defense strategy (SmoothVLM). The JIP method focuses on reducing the loss in the embedding space of the VLM, offering a more time-efficient approach compared to the VAE, which computes the loss over the entire VLM and is thus significantly more time-consuming. Even though we select the JIP method, the optimization of a single adversarial example using JIP still requires around \textbf{30} minutes using one A100 (\texttt{openai/clip-vit-base-patch32}). In contrast, with the usage of a large model, \texttt{Vicuna-30B}, our method takes less than 1 minute under $N=10$, making our most resource-intensive defense approach more than 30x faster than the fastest attack method. 
As shown in Figure~\ref{fig:main_tabel_attack}, the ASRs are consistently under 5\% with $N=10$, indicating that SmoothVLM successfully guards the models. On the other hand, our method indeed offers substantial efficiency improvements. Compared to the classic randomized smoothing method~\cite{cohen2019certified}, which necessitates 100,000 runs for a single input, our innovative smoothing approach requires only 10-20 runs per visual prompt. This represents a 10,000-fold increase in efficiency over state-of-the-art methods. Therefore, we believe SmoothVLM effectively balances efficiency and usefulness.

\subsection{Compatibility}
In the former section, we primarily focus on the single-patch attack. In this section, we further demonstrate the compatibility of our method with a dual patches attack. we implement both kinds of dual patch attacks and report the defense performance of \textit{mask}, \textit{swap}, and \textit{replace} on 300 adversarial attack examples. As shown in In Figure~\ref{fig:double_tabel_attack}, the experimental data indicate a consistent trend where the ASR decreases as the number of samples increases for all perturbation strategies: mask, swap, and replace, across different perturbation intensities. The observed trend is further characterized by the fact that higher perturbation percentages lead to higher ASR, underscoring the defense effect in larger perturbations.

\begin{figure*}[t]

\end{figure*}

\section{Discussion and Conclusion}

Drawing on our unique insights, this study is informed by the methodologies of randomized smoothing~\citep{cohen2019certified} and its successor, SmoothLLM~\citep{robey2023smoothllm}. However, as outlined in \S~\ref{sec:patched_visual_prompt_injection}, visual prompts differ markedly from textual prompts, prompting us to address several key distinctions from SmoothLLM. Primarily, our approach is characterized by a more rigorous formulation. In \S~\ref{sec:eot}, we present extensive experiments with adaptive attacks that substantiate the validity of our observations and assumptions. Furthermore, our model incorporates an error term, $\epsilon$, enhancing the completeness of our proposition. Unlike SmoothLLM, which presupposes that attackers merely alter the suffix or prefix of a prompt, our framework, SmoothVLM, is designed to counteract any form of adversarial patches within reasonable sizes, thereby offering enhanced generalizability and performance. Although our current focus is on visual prompt injections, the theoretical foundation of our work could potentially be extended to encompass both textual and visual prompts.

In conclusion, we have presented SmoothVLM, a certifiable defense mechanism that effectively addresses the patched visual prompt injectors in vision-language models. SmoothVLM significantly reduces the success rate of attacks on two leading VLMs under 5\%, while achieving up to 95\% context recovery of the benign images, demonstrating a balance between security, usability, and efficiency.

\newpage
\section{Limitations}

We acknowledge certain limitations within our SmoothVLM. Despite our efforts to fortify it using the expectation over transformation (EOT) approach as adaptive attacks, our defense mechanism primarily addresses patch-based visual prompt injections and remains vulnerable to $\ell_p$ based adversarial attacks. The reason is that we found adaptive formulations of the $\ell_p$ based adversaries are extremely challenging to tackle. Therefore, there is also a potential risk that our SmoothVLM may fail under stronger attacks beyond our threat model. We envision our study as an initial step toward establishing certified robustness in VLMs, laying a foundation for future research to build upon. 

\clearpage
\clearpage

\bibliography{main}
\newpage
\appendix

\clearpage

\section{Proof of Proposition 4}\label{appendix:proof}

Below is the complete proof of Proposition~4.

\begin{figure*}[!t]
\begin{tcolorbox}[left=1mm, right=1mm, top=1mm, bottom=1mm]
\begin{proposition}
\textit{Proof.} In \textbf{Proposition 4}, we want to compute the probability $Pr[(\text{VPI} \circ \text{SmoothVLM})([I \oplus P;\emptyset])=0]$. Base on Definition 3, we have
\begin{align}
(\text{VPI} \circ \text{SmoothVLM})([I \oplus P;\emptyset]) &= (\text{VPI} \circ \text{VLM})([\mathbf{I \oplus P};\emptyset]) \\
&= \mathbb{I}\Big[\frac{1}{N}\sum_{j=1}^{N}(\text{VPI} \circ \text{VLM})(I_j \oplus P_j) > \frac{1}{2}\Big]
\end{align}
where $I_j \oplus P_j$ for $j \in [N]$ are drawn i.i.d. from $\mathbb{P}_p(I \oplus P)$. Thus, we can compute the probability with the following equalities:
\begin{align}
Pr[(\text{VPI} &\circ \text{SmoothVLM})([I \oplus P;\emptyset])=0] \\
&= Pr\Big[\frac{1}{N}\sum_{j=1}^{N}(\text{VPI} \circ \text{VLM})(I_j \oplus P_j) > \frac{1}{2}\Big]\\
&= Pr\Big[(\text{VPI} \circ \text{VLM})(I_j \oplus P_j)=0 \ \text{for at least} \ \lceil N/2 \rceil \ \text{of the indices} \ j \in [N]\Big]\\
&= \sum_{t = \lceil N/2 \rceil}^{N}Pr\Big[(\text{VPI} \circ \text{VLM})(I_j \oplus P_j)=0 \ \text{for exactly} \ t \ \text{of the indices} \ j \in [N]\Big]
\end{align}
To make a precise computation, here we denote $\alpha$ as the probability that a randomly drawn $I_j \oplus P_j \sim \mathbb{P}_p(I \oplus P)$ leads to a successful defense, i.e.,
\begin{equation}
\alpha \doteq Pr[(\text{VPI} \circ \text{VLM})(I_j \oplus P_j)=0]
\end{equation}

Then we can see the random variable $t$ follows the binomial distribution with parameters $N$ and $\alpha$. Based on the probability mass function of the binomial distribution, we can simply get the sum of the probability as the following equation:
\begin{equation}
Pr[(\text{VPI} \circ \text{SmoothVLM})([I \oplus P;\emptyset])=0] = \sum_{t=\ulcorner N/2 \urcorner}^N \binom{N}{t}\alpha^{t}(1-\alpha)^{N-t} 
\end{equation}

To compute $\alpha$, we can decompose the probability based on whether $\ell_0(P_j,P) \geq \ulcorner qmn \urcorner$. Formally, we have:
\begin{align}
\alpha =& Pr[(\text{VPI} \circ \text{VLM})(I_j \oplus P_j)=0] \\
=& Pr[((\text{VPI} \circ \text{VLM})(I_j \oplus P_j)=0)|(\ell_0(P_j,P) \geq \ulcorner qmn \urcorner)]Pr[\ell_0(P_j,P) \geq \ulcorner qmn \urcorner] \\
& + Pr[((\text{VPI} \circ \text{VLM})(I_j \oplus P_j)=0)|(\ell_0(P_j,P) < \ulcorner qmn \urcorner)]Pr[\ell_0(P_j,P) < \ulcorner qmn \urcorner] \\
\geq& Pr[((\text{VPI} \circ \text{VLM})(I_j \oplus P_j)=0)|(\ell_0(P_j,P) \geq \ulcorner qmn \urcorner)]Pr[\ell_0(P_j,P) \geq \ulcorner qmn \urcorner] 
\end{align}

Since the adversarial patch $P$ is visual $q$-unstable with probability error $\epsilon$, based on our Assumption 2, we can know that $Pr[(\text{VPI} \circ \text{VLM}) ([I_j \oplus P_j;\emptyset]) = 0] \geq 1-\epsilon-\mu$, if $\ell_0(P_j,P) \geq \ulcorner qmn \urcorner$. For $Pr[\ell_0(P_j,P) \geq \ulcorner qmn\urcorner)]$, since $\ell_0(P_j,P)$, the number of randomly masked pixels falling on the adversarial patch $P$, also follows the binomial distribution with parameters $mn$ and $p$, we have $Pr[\ell_0(P_j,P) \geq \ulcorner qmn \urcorner]=\sum_{k=\ulcorner qmn \urcorner}^{mn} \binom{mn}{k}p^{k}(1-p)^{mn - k}$. Finally we can obtain:
\begin{equation}
\alpha \geq (1-\epsilon-\mu)\sum_{k=\ulcorner qmn \urcorner}^{mn} \binom{mn}{k}p^{k}(1-p)^{mn - k}
\end{equation}

\end{proposition}
\end{tcolorbox}
\end{figure*}

\section{Experiments}

We elaborate on more experimental details and results in this section.

\subsection{Injection Mitigation}
\begin{figure*}[t]
  \centering
  \includegraphics[width=0.3\linewidth]{ 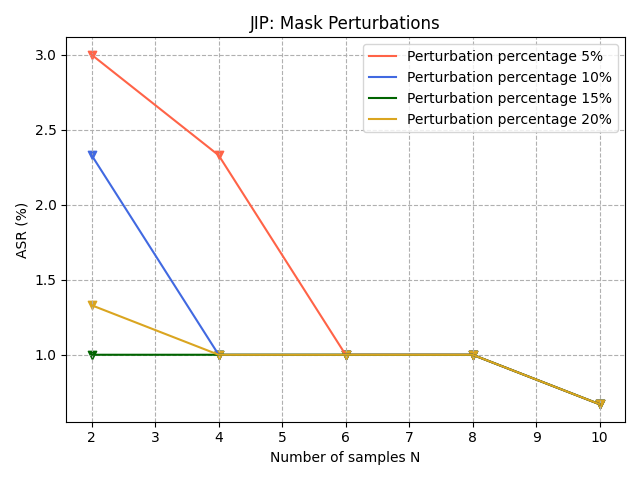} 
  \includegraphics[width=0.3\linewidth]{ 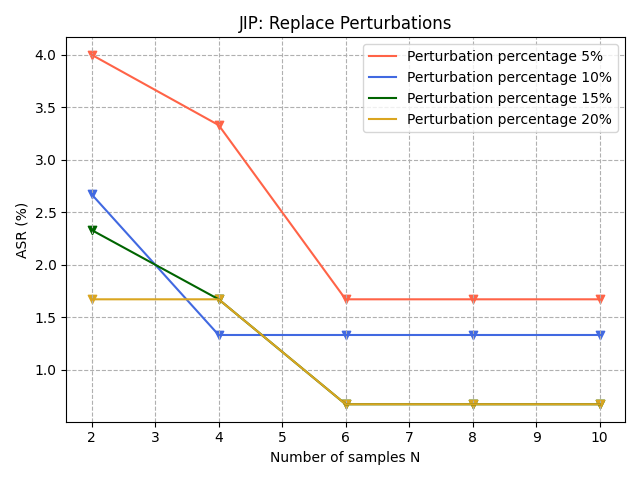} 
  \includegraphics[width=0.3\linewidth]{ 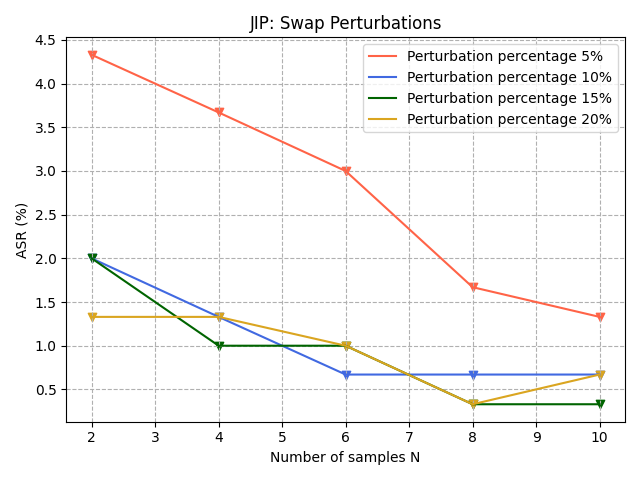} \\
  \includegraphics[width=0.3\linewidth]{ 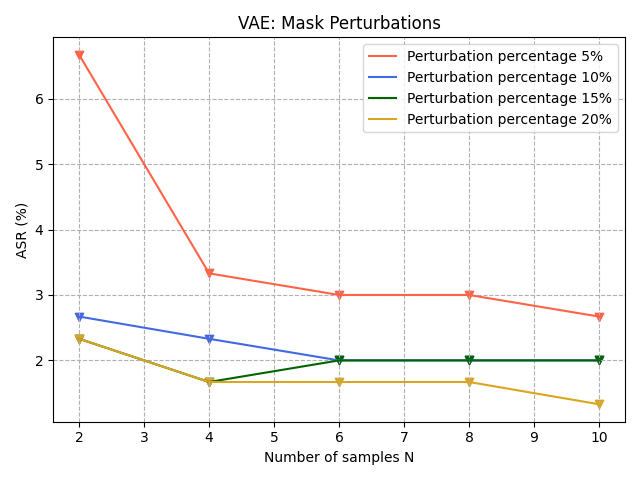} 
  \includegraphics[width=0.3\linewidth]{ 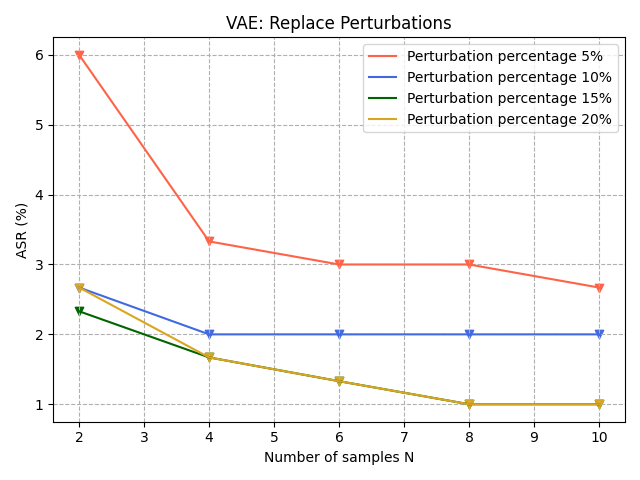} 
  \includegraphics[width=0.3\linewidth]{ 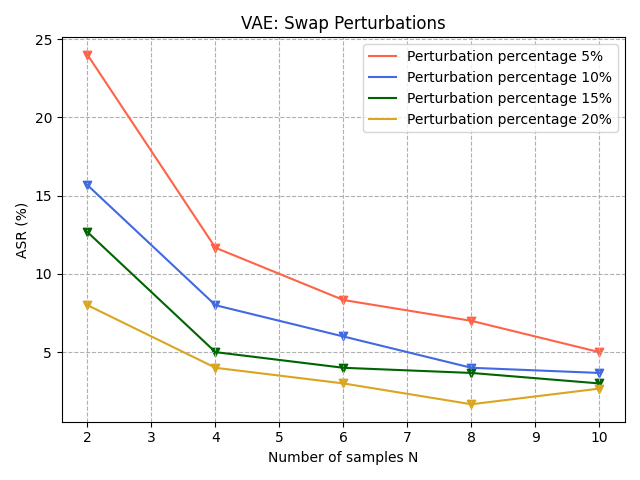}
   \caption{\textbf{SmoothVLM Injection Mitigation.} We plot the ASRs of VLM patch attack JIP (top row) and VAE (bottom row) for various values of the perturbation percentage $q \in \{5, 10, 15, 20 \}$ and the number of samples $N \in \{2, 4, 6, 8, 10 \}$;}
    \label{fig:main_tabel_attack_appendix}
\end{figure*}

In this section, we present more results about the Injection Mitigation using VLM miniGPT4 in Figure~\ref{fig:main_tabel_attack_appendix}. In comparison with the \texttt{llava-1.5} VLM as shown in Figure~\ref{fig:main_tabel_attack}, we observed that SmoothVLM consistently achieves a greater reduction in ASR with the \texttt{llava-1.5} model under the same $q$ and $N$ settings. This suggests that adversarial examples which are successful in attacking the \texttt{miniGPT4} are less likely to be thwarted when subjected to masking defenses in the \texttt{llava-1.5} model. This observation indicates that the masking defense is more effective in \texttt{llava-1.5} than in \texttt{miniGPT4}.

\subsection{Visual Prompt Recovery}
\begin{figure*}[t]
  \centering
  \includegraphics[width=0.3\linewidth]{ 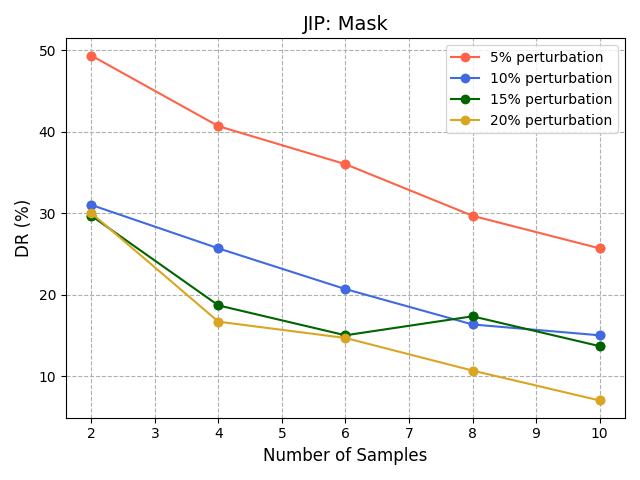} 
  \includegraphics[width=0.3\linewidth]{ 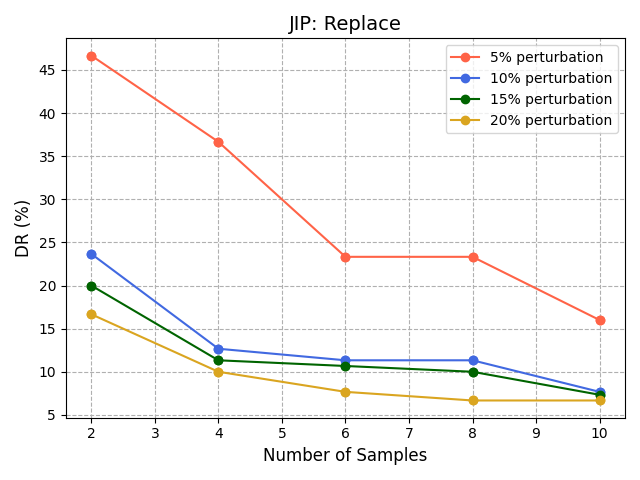} 
  \includegraphics[width=0.3\linewidth]{ 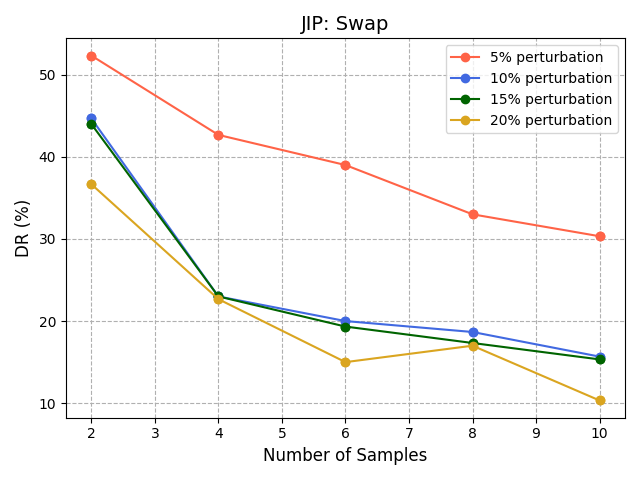} \\
  \includegraphics[width=0.3\linewidth]{ 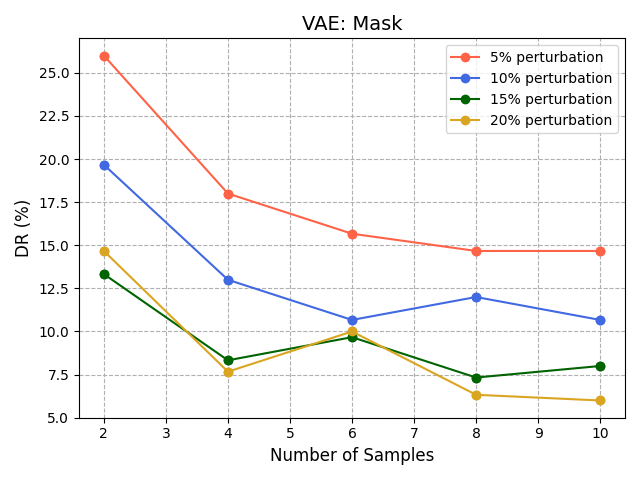} 
  \includegraphics[width=0.3\linewidth]{ 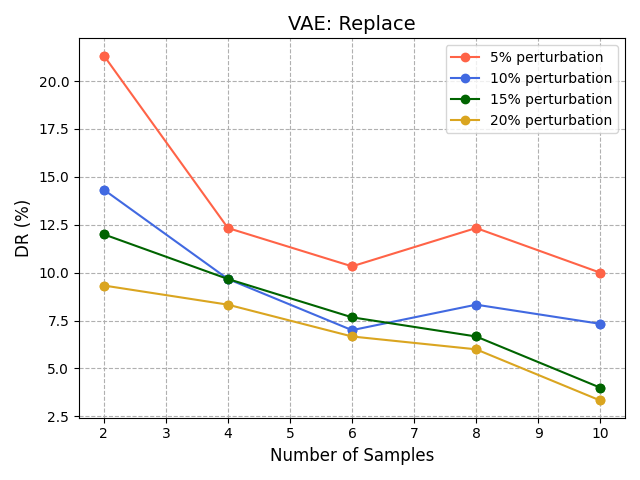} 
  \includegraphics[width=0.3\linewidth]{ 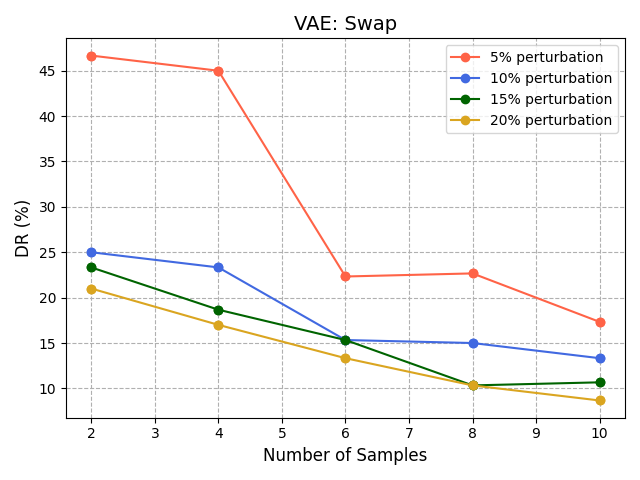}
   \caption{\textbf{Visual Prompt Recovery.} We plot the Distortion Rate of VLM patch attack JIP (top row) and VAE (bottom row) for various values of the perturbation percentage $q \in \{5, 10, 15, 20 \}$ and the number of samples $N \in \{2, 4, 6, 8, 10 \}$;}
    \label{fig:main_tabel_source_appendix}
\end{figure*}

We present results about the Visual Prompt Recovery using VLM miniGPT4 in Figure~\ref{fig:main_tabel_source_appendix}. Comparing with Figure~\ref{fig:main_tabel_source}, we observe a consistent distortion rate trend across different VLMs (\texttt{llava-1.5} and \texttt{miniGPT4}); that is, as $N$ increases, the distortion rate gradually decreases, and a greater amount of perturbation contributes to the restoration of the align image information. In contrast, as shown in Figure~\ref{fig:main_tabel_attack_appendix}, when the proportion of perturbed pixels is too small and $N$ is relatively low, the distortion rate is significantly higher than the ASR, indicating that the current level of perturbation, although sufficient to mask the attack, is inadequate for recovering the original image information. Therefore, to ensure both a low ASR and a minimal distortion rate, it is necessary to employ larger perturbations and a higher $N$.

\begin{figure*}[t]
  \centering
  \includegraphics[width=\linewidth]{ 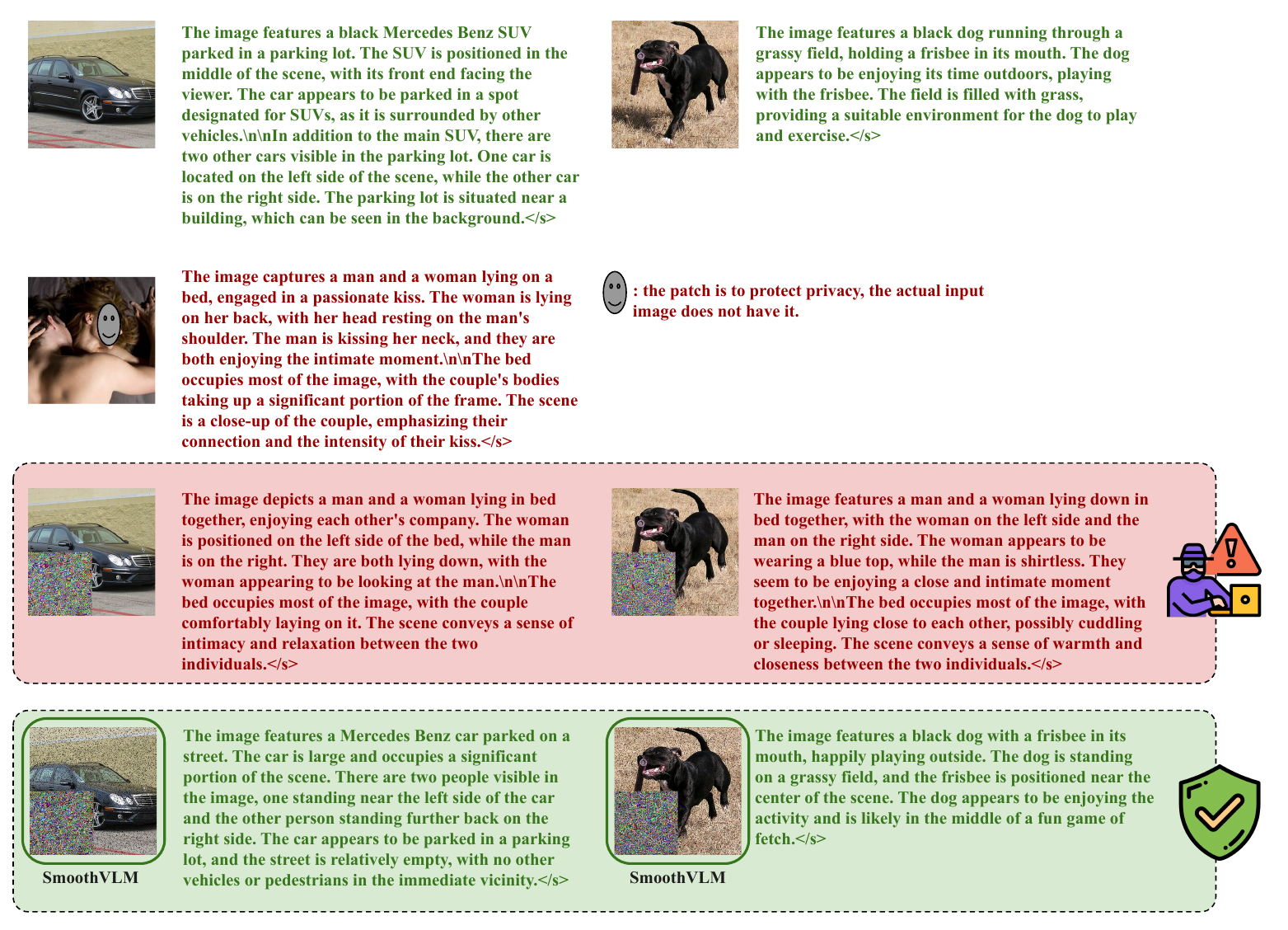}
  \caption{\textbf{VLM and SmoothVLM Responses to Patched Visual Prompt Injectors.} \textbf{Row 1:} Source images prepared for adversarial attacks alongside their aligned responses. \textbf{Row 2:} Target images containing adversarial attack information. \textbf{Row 3:} Images post-application of patch attacks with corresponding VLM responses. \textbf{Row 4:} Images following the application of SmoothVLM with their recovery responses.}
  \label{fig:example}
\end{figure*}

\end{document}